\documentclass[journal]{IEEEtran} 
\IEEEoverridecommandlockouts
\usepackage{booktabs} 
\usepackage{tabularx} 
\usepackage{xcolor} 
\usepackage{graphicx} 
\usepackage{cite}
\usepackage{amsmath,amssymb,amsfonts,amsthm}
\usepackage{threeparttable}
\usepackage{graphicx}
\usepackage{textcomp}{\tiny }
\usepackage{multirow}
\usepackage{xcolor}
\usepackage{bm,color}
\usepackage{lmodern} 
\usepackage{makecell}
\usepackage[citecolor=red,urlcolor=red]{hyperref}
\hypersetup{
	colorlinks=true,
	linkcolor=blue,
	filecolor=red,      
	urlcolor=blue,
	citecolor=blue,
}
\usepackage{graphicx} 
\usepackage{subfigure}
\usepackage{colortbl}
\usepackage{hyperref}
\usepackage{nomencl}
\makenomenclature

\def\BibTeX{{\rm B\kern-.05em{\sc i\kern-.025em b}\kern-.08em
		T\kern-.1667em\lower.7ex\hbox{i}\kern-.125emX}}

\usepackage{booktabs}
\usepackage{bm}
\usepackage{algorithm}
\usepackage{algpseudocode}
\usepackage{amsmath}

\begin{document}
	\title{ \textcolor{black} { A Self-Rotating Tri-Rotor UAV for  \\
			 Field of View Expansion and Autonomous Flight}}
		 
		 	\author{
		 	\IEEEauthorblockN{Xiaobin Zhou}, 
		 	\IEEEauthorblockN{Zihao Zheng}, 
		 	\IEEEauthorblockN{Aoxu Jin},
		 	\IEEEauthorblockN{Lei Qiang},
		 	\IEEEauthorblockN{Bo Zhu\IEEEauthorrefmark{3*}}

		 	
		 	\thanks{
		 			 		This work was supported in part by the National Natural Science Foundation of China under Grant 62303412, in part by  Fundamental and Interdisciplinary Disciplines Breakthrough Plan of the Ministry of Education of China JYB2025XDXM902, in part by the Open Fund of State Key Laboratory of Digital Intelligent Technology for Unmanned Coal Mining RAO2026K09， and in part by the Natural Science Foundation of Zhejiang Province Huzhou City under Grant 2023YZ01. (Corresponding author: Bo Zhu)
		 	
		 	The authors are with the School of Robotics and Automation, Nanjing University, Suzhou, 215163, China. (E-mail: xiaobin\_nju@nju.edu.cn, zzh211106370@gmail.com, oliverking0219@gmail.com, qianglei1204@gmail.com, zhubo@nju.edu.cn.) 		 	}
		 	
		 }	
	\maketitle
\markboth{2026 IEEE International Conference on Robotics and Automation (ICRA). June 1-5, 2026, Vienna, Austria}%
{Shell \MakeLowercase{\textit{et al.}}:   }
\maketitle	

\begin{abstract}
	Unmanned Aerial Vehicles (UAVs) perception relies on onboard sensors like cameras and LiDAR, which are limited by the narrow field of view (FoV). We present Self-Perception INertial Navigation Enabled Rotorcraft (SPINNER), a self-rotating tri-rotor UAV for the FoV expansion and autonomous flight. Without adding extra sensors or energy consumption, SPINNER significantly expands the FoV of onboard camera and LiDAR sensors through continuous spin motion, thereby enhancing environmental perception efficiency. SPINNER  achieves full 3-dimensional position and roll--pitch attitude control using only three brushless motors, while adjusting the rotation speed via anti-torque plates design. To address the strong coupling, severe nonlinearity, and complex disturbances induced by spinning flight, we develop a disturbance compensation control framework that combines nonlinear model predictive control (MPC) with incremental nonlinear dynamic inversion. Experimental results demonstrate that SPINNER maintains robust flight under wind disturbances up to 4.8 \,m/s and achieves high-precision trajectory tracking at a maximum speed of 2.0\,m/s. Moreover, tests in parking garages and forests show that the rotational perception mechanism substantially improves FoV coverage and enhances perception capability of SPINNER.
\end{abstract}


\IEEEpeerreviewmaketitle

\section{Introduction}

\IEEEPARstart{M}{iniature}  unmanned aerial vehicles (UAVs) have been widely deployed in tasks such as search and rescue, cave exploration, and aerial mapping \cite{mellinger2011minimum, Philipp2022Agilicious}. Their autonomous capabilities typically rely on onboard sensors, such as cameras and LiDARs, to achieve localization, mapping, and obstacle avoidance. However, most existing sensors suffer from limited field of view (FoV), which significantly constrains perception efficiency and flight safety of UAV~\cite{Wang2024Multi}. 

\begin{figure*}
	\centering
	\includegraphics[width=0.950\linewidth]{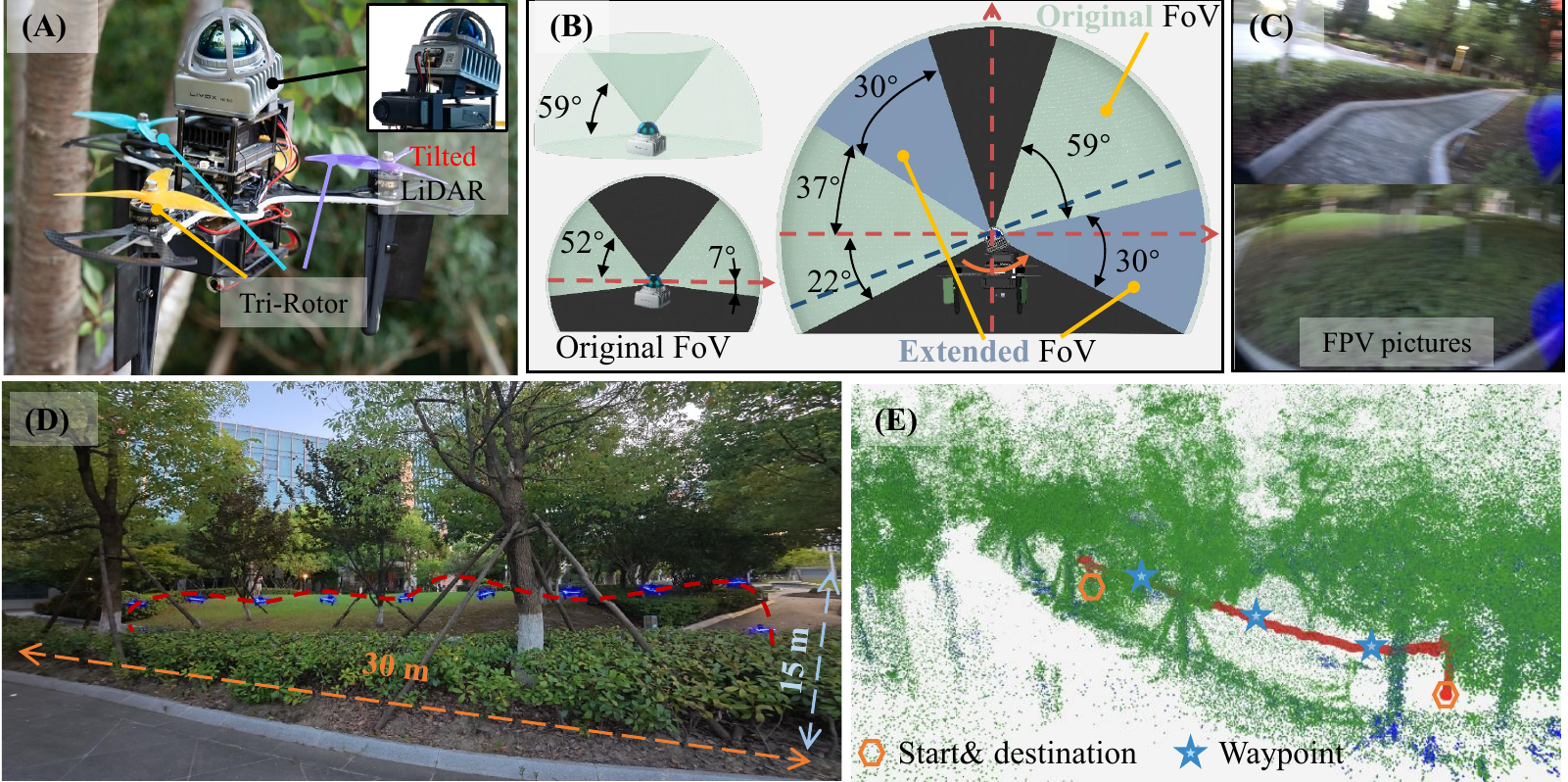}
	\caption{\textcolor{black}{ Autonomous waypoint navigation of SPINNER in a forested environment. (A) SPINNER is equipped with three rotors and motors, a first-person view (FPV) camera, and a tilted LiDAR (Mid 360). (B) Through self-rotation, the vertical FoV is extended from $59^\circ$ to $139^\circ$. (C) The FPV camera captures a wider visual scene due to the UAV's self-rotation. (D)–(E) The real flight process and the corresponding 3D point cloud map in the forest. The flight trajectory is shown as a red path. The video is available at \url{https://hari-robotics.github.io/SPINNER-ICRA-2026/}.}} 
	\label{forest}
\end{figure*}

Although some studies attempt to mitigate the perceptual challenge of UAV posed by narrow FoV, expanding the FoV remains an effective approach to improving task efficiency~\cite{chen2023self}. Current strategies for FoV extension mainly fall into two categories: the use of wide-angle sensors and multi-sensor fusion. Nevertheless, wide-angle sensors often introduce image distortion, limited vertical FoV, or poor resolution~\cite{Gurtner2009Investigation, Gao2020Autonomous}. On the other hand, multi-sensor fusion approaches typically result in increased system cost, weight, and power consumption, making them unsuitable for small-sized UAV platforms with strict payload and energy constraints~\cite{Karimi2021LoLa, Liu2024OmniNxt}.

An optional approach to expanding sensor FoV is to exploit the UAV’s inherent counter-torque and its capability for self-induced rotational motion. In conventional multirotor UAV designs, counter-torque is typically neutralized by configuring adjacent brushless motors to rotate in opposite directions, ensuring near-zero net yaw torque during hover. However, if properly utilized, counter-torque can be harnessed to induce controlled UAV body rotation, thereby expanding the sensor FoV without compromising flight safety. This method avoids the need for additional sensing hardware while enhancing perception capability, making it beneficial for complex flight missions. For example, the studies in~\cite{ chen2023self} demonstrate the successful application of rotational motion of a UAV to deal with FoV constraints.

Nevertheless, the high-speed and intrinsic spinning motion of UAVs introduces significant challenges in vehicle modelling, control, and navigation.  From the perspectives of modelling and control, continuous high-speed rotation significantly alters the UAV aerodynamics, introducing strong coupling, severe nonlinearity, and complex disturbances~\cite{Nan2022Nonlinear}. Furthermore, conventional multirotor UAVs rely on the counter-torque from opposing motors to maintain stable yaw control, and sustained spinning may lead to thrust saturation in individual motors~\cite{zhou2024internal, zhou2025rotor}. As a result, traditional linear control strategies may suffer from degraded performance. It must be replaced by nonlinear disturbance compensation control algorithms, capable of mitigating the model mismatches and considering thrust limits.  From the perspectives of perception and navigation, high-speed rotation causes harsh motion blur and rapid FoV shifts. The onboard IMU tends to exhibit significant measurement errors under high angular velocities, which physically undermines the reliability of visual sensors~\cite{Sun2021Autonomous}. The low inter-frame image overlap and scan distortions severely impact the performance of classical visual-inertial odometry, LiDAR-inertial odometry, and SLAM systems, which heavily depend on feature matching and temporal continuity~\cite{he2023point}.

To address the aforementioned challenges in UAV modelling, control, and navigation, while simultaneously achieving the goal of FoV expansion, we propose a  self-rotating UAV, SPINNER (Self-Perception INertial Navigation Enabled Rotorcraft). As illustrated in Fig.~\ref{forest},  SPINNER employs three common brushless motors as actuators to control its full 3D position and attitude. During normal hovering, the three motors generate equal thrust to counteract gravity and maintain roll and pitch moment balance. Meanwhile, the counter-torque naturally induces body rotation, which expands the effective FoV of the onboard LiDAR and camera, without requiring any additional actuators. Compared to conventional quadrotors, SPINNER features a simpler mechanical structure with fewer actuators, resulting in reduced overall weight and lower power consumption. The main contributions of this work are summarized as follows:

\begin{itemize}
	
	\item  A  self-rotation UAV, SPINNER, is proposed, equipped with both LiDAR and camera, where the rotation enables expanding the FoV of both sensors  (see Section ~\ref{Preliminary}). Additionally, the adjustment of the UAV rotational speed or yaw rate is achieved by designing the width of anti-torque plates under each rotor.
	
	\item A nonlinear disturbance compensation control framework is developed based on nonlinear model predictive control (MPC) and incremental nonlinear dynamic inversion (INDI) (see Section ~\ref{sec:dynamics_control}) for SPINNER. Experimental results demonstrate high-precision trajectory tracking at speeds up to 2.0\,m/s. The system also exhibits strong robustness to external disturbances such as 4.8\,m/s gusts.
	
	\item Extensive tests in parking garages and forests show that SPINNER can fully rely on its onboard sensors and computation to perform autonomous navigation, without requiring any external infrastructure or human intervention (see Section~\ref{ Real-World Experiments}). 
	
\end{itemize}

The remainder of this paper is organized as follows. Related work is reviewed in Section~\ref{Related Work}. Sections~\ref{Preliminary} and \ref{sec:dynamics_control} present the UAV design, system dynamics and the controller design.  Real-world experimental results are provided in Section~\ref{ Real-World Experiments}. Finally, concluding remarks are presented in Section~\ref{Conclusion}.

\section{Related Work}
\label{Related Work}

\subsection{Design and Control of Self-Rotating UAVs }

According to their actuation principles and structural configurations, existing actively self-rotating UAVs include the following types: seed-inspired self-rotating drones, conventional multirotors, rotorcraft without swashplate mechanisms, and swashplate-based aerial vehicles. Seed-inspired self-rotating UAVs, as explored in~\cite{bhardwaj2022design,cai2023self,cai2023modeling}, mimic the flight dynamics of maple seeds and combine the benefits of fixed-wing and multirotor platforms. They generate lift through rotating blade surfaces rather than direct propeller thrust, but require customized blade designs and involve complex aerodynamics, making controller design challenging.   In~\cite{chen2023self}, a swashplate-free UAV achieves full 3D position control by modulating motor speeds, while blade pitch is passively adjusted via tilted hinges. This configuration enables desirable agility without active attitude modulation, but imposes mechanical stress on bearings and blades, potentially reducing system durability.  The single-rotor UAV in~\cite{zhang2016controllable} features an extremely simplified mechanical structure with only one actuator. A cascaded control scheme is adopted, comprising an inner-loop linear quadratic regulator for attitude stabilization and an outer-loop position controller. Robustness is evaluated through disturbance rejection and input saturation probability, and experiments demonstrate stable hovering even under perturbations such as mid-air throws. In~\cite{xu2025aerial}, a dual-mode reconfigurable aerial robot is introduced, capable of switching between conventional quadrotor and self-rotating flight modes. The platform achieves decoupled control of position and roll/pitch attitude, and shows promising performance in tasks such as mapping and navigation in confined spaces. However, the implementations of~\cite{zhang2016controllable,xu2025aerial} remain limited to hovering or low-speed flight.  In summary, the need for customized electromechanical components or the restriction to low-speed flight pose challenges for the applicability of the above UAVs in real-world scenarios.

\subsection{ Autonomous Flight under Self-Rotation }
Autonomous navigation is critical for UAVs operating in unknown and GNSS-denied environments. However, self-rotating UAVs face major challenges in vision-based navigation due to severe motion blur and rapid FoV transitions caused by high-speed body rotation. Consequently, most existing self-rotating UAV platforms lack the capability for fully autonomous navigation in unknown environments. The works in~\cite{bhardwaj2022design,cai2023self,cai2023modeling} primarily emphasize electromechanical design and control. Although full-state estimation is performed, these studies rely on external motion capture systems for position and attitude feedback, restricting their use to controlled indoor environments. Similarly, other efforts depend on GPS or RF-based positioning for state estimation, which limits applicability in GNSS-denied settings. The work in ~\cite{Sun2021Autonomous} estimate the full state of a self-rotating quadrotor using a downward-facing camera (standard or event-based). Moreover, beyond state estimation, most self-rotating UAVs~\cite{win2021agile,sharp2016micro,isaacs2014gps,fregene2010development} lack onboard 3D mapping capabilities, which are essential for autonomous navigation in unknown environments and for exploiting extended FoV. The UAV in~\cite{chen2023self} is closely related to this work, demonstrating  FoV expansion and excellent autonomous navigation without external infrastructure. However, it focuses on extending the sensor’s horizontal FoV to $360^\circ$ while neglecting vertical perception coverage.

\definecolor{mygray}{gray}{.9}
\begin{table}[!t]
	\renewcommand{\arraystretch}{1.5}
	\caption{Component Configuration of \textsc{SPINNER}}
	\label{tab:hardware}
	\centering
	\resizebox{\columnwidth}{!}{%
		\begin{tabular}{llll}
			\toprule
			\textbf{Component} & \textbf{Model} & \textbf{Units} & \textbf{Weight (g)} \\
			\midrule
				\rowcolor{mygray}
			\rowcolors{2}{mygray}{} 
			LiDAR              & Mid 360                  & 1 & 277 \\
			
			Battery            & TATTU FPV 2300mAh 4S     & 1 & 236 \\
				\rowcolor{mygray}
			Brushless Motor    & T-Motor F60Pro           & 3 & 134 \\
			Onboard Computer   & NVIDIA Jetson Orin NX    & 1 & 120 \\
				\rowcolor{mygray}
			Anti-Torque Plate       & 3D Printed Parts         & 3 & 84  \\
			Camera             & Hawk eye camera   & 1 & 38  \\
				\rowcolor{mygray}
			ESC                & HAKRC 65A                & 1 & 14  \\
			Propellers         & Gemfan 51477             & 3 & 12  \\
				\rowcolor{mygray}
			Flight Controller  & Holybro Kakute H7 V1     & 1 & 9   \\
			\bottomrule
		\end{tabular}%
	}
\end{table}

\begin{figure}
	\centering
	\includegraphics[width=0.850\linewidth]{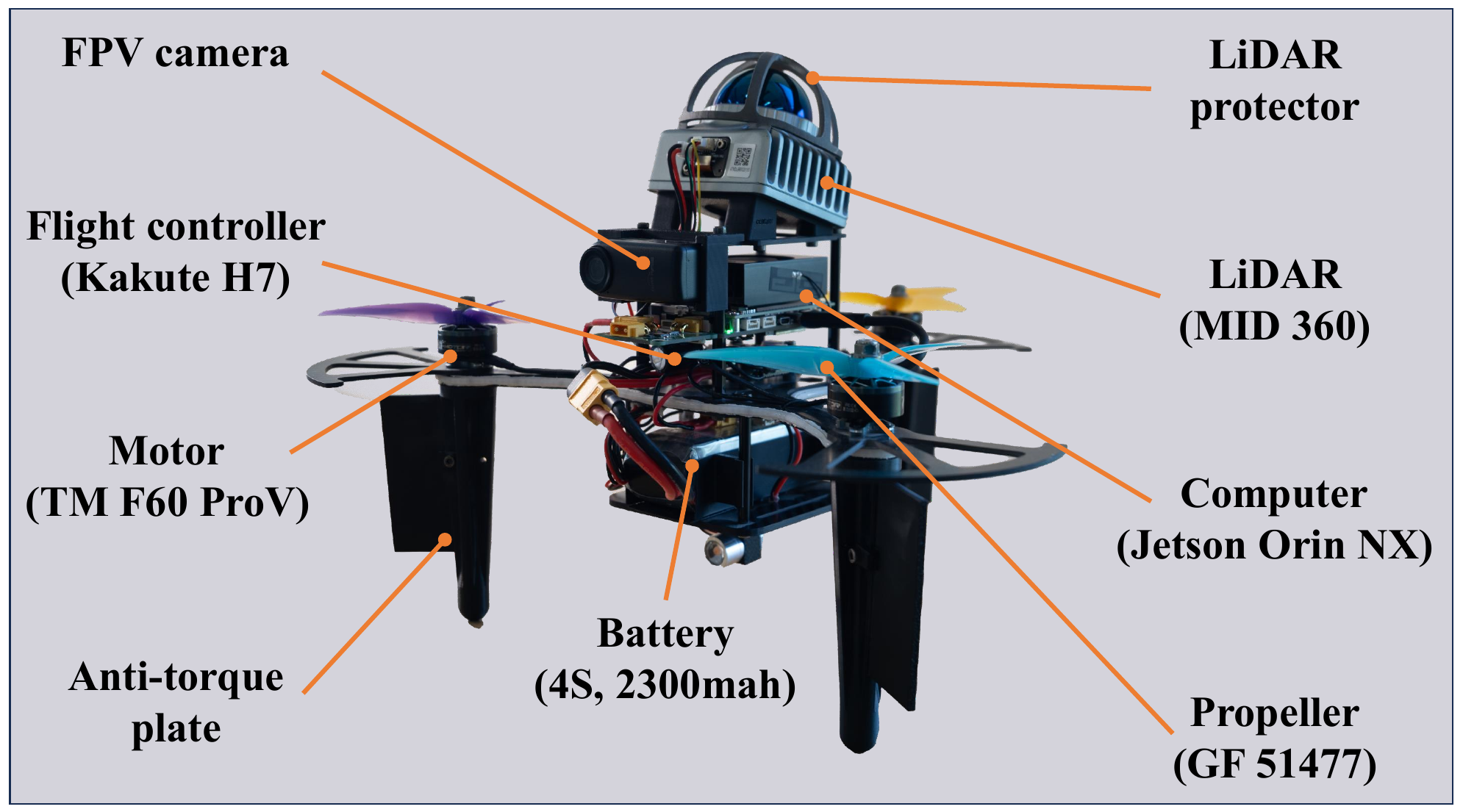}
	\caption{\textcolor{black}{ Overview of the UAV platform and components. }}
	\label{hardware}
\end{figure}

\begin{figure}
	\centering
	\includegraphics[width=0.990\linewidth]{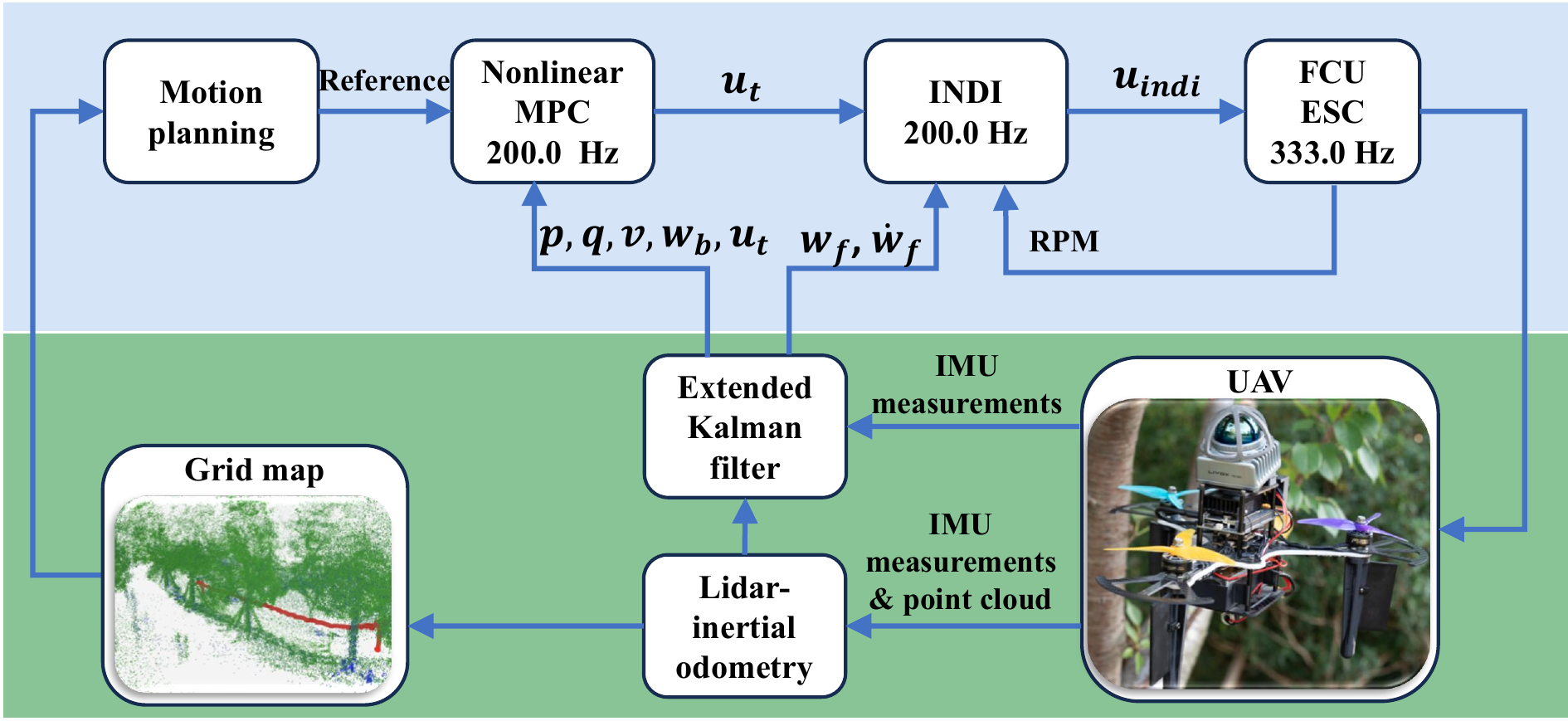}
	\caption{\textcolor{black}{The developed software system in real-world experiment. }}
	\label{flowchart}
\end{figure}

\section{ UAV Design }
\label{Preliminary}

\subsection{Tri-rotor UAV Platform Design} 
All experiments are conducted on a custom-designed and assembled tri-rotor UAV platform. The hardware specifications are summarized in Fig.~\ref{hardware}. The total takeoff weight is  1150\,$g$, and the overall dimensions are 31.6\,$cm$ × 31.6\,$cm$ × 18.2\,$cm$. The SPINNER integrates a propulsion system comprising propellers, brushless motors, a flight controller, and a battery; a perception system including an onboard computer, a visual sensor, and a LiDAR module; and several auxiliary components such as the anti-torque plates and carbon fibre frame. Table \ref{tab:hardware} lists the key components with model specifications of SPINNER.

The imbalance in yaw torque causes SPINNER to enter a high-speed spinning motion accompanied by intense vibrations. Under high-speed spinning, the efficiency of LiDAR-based odometry may deteriorate. The spin motion of SPINNER originates from the counter-torque produced by the rotating propellers, with the yaw torque expressed as: 
\begin{equation} 
	\tau_{z} = \left[-k_y, \ k_y, \ k_y\right] \left[  u_0, \ u_1, \ u_2\right]^{T}, 
\end{equation}
where $\tau_{z}$ denotes the yaw torque and $k_y$ is the coefficient of counter-torque. $u_0$, $u_1$, and $u_2$ are the thrusts.  The imbalance in $\tau_{z}$ leads to an increase in the spin speed until the air resistance balances $\tau_{z}$. To mitigate excessive spin speed, a viable strategy is to adjust the air resistance. Therefore, anti-torque plates are designed under the rotors, which increase air resistance and compensate for the motor's counter-torque. The detail can be referred to section \ref{Anti-torque vanes design}.

\subsection{Software Framework}

To support autonomous flight, a lightweight software stack (see Fig.~\ref{flowchart}) runs on the onboard NVIDIA Jetson Orin NX. Communication with the flight controller uses MAVROS~\cite{Mavros2024git}. The system includes autonomous perception, onboard planning, high-level control, and low-level actuation. The FAST-LIVO2 algorithm \cite{Zheng2025FAST}, an efficient LiDAR-inertial odometry framework, is utilized. To obtain a spatial-temporal trajectory from the given waypoints, the trajectory representation  is applied with MINCO \cite{Wang2022Zhe}. To follow the reference trajectory, nonlinear MPC and INDI modules compute desired revolutions per minute (RPM) command of motors at 200\,Hz. RPM is sent to the Kakute H7 V1.3 flight control unit (FCU), which relays DShot signals to a Tekko 32 F4 electronic speed control (ESC). ESC adjusts motor speed and sends measured RPM back to INDI for closed-loop control. 

As illustrated in Fig.~\ref{forest}(B), the vertical FoV of the Livox Mid-360 LiDAR is expanded by leveraging SPINNER’s continuous self-rotation. While the LiDAR natively provides a vertical FoV of  $59^\circ$, the LiDAR is mounted with a fixed tilt angle $15^\circ$ on the tri-rotor platform (see Fig.~\ref{forest}(A)). This tilt, combined with rapid body rotation, causes the original FoV to sweep across a wider vertical region.  As shown in Fig.~\ref{forest}(B), this motion results in an effective FoV extension of $30^\circ$, yielding a total vertical coverage of up to $89^\circ$. The FPV camera is mounted adjacent to the LiDAR (see Fig.~\ref{forest}(C)), which captures a broader horizontal view of the flight environment. This extended FoV enables the onboard sensor to perceive obstacles and terrain, which is especially beneficial for navigation in cluttered or GNSS-denied environments such as forests and caves.

\newtheorem{remark}{Remark}

\section{Dynamics and Control}
\label{sec:dynamics_control}

\subsection{Tri-Rotor UAV Dynamics}

The tri-rotor UAV operates within two right-handed coordinate systems: the inertial frame \( \mathcal{F}_w: \{\bm{x}_w, \bm{y}_w, \bm{z}_w\} \), with \(\bm{z}_w\) oriented upward (opposing gravity), and the body frame \( \mathcal{F}_b: \{\bm{x}_b, \bm{y}_b, \bm{z}_b\} \), where \(\bm{x}_b\) points forward and \(\bm{z}_b\) aligns with the thrust direction. The rotation from \( \mathcal{F}_w \) to \( \mathcal{F}_b \) is described by the rotation matrix \( \bm{R}_{b} \), parameterized by a quaternion \( \bm{q} = [q_0, q_1, q_2, q_3]^T \).

The UAV is modeled as a rigid body with the following translational and rotational dynamics:

\begin{equation}
	m\ddot{\bm{p}} = T \bm{z}_b - \bm{R}_{b}\bm{D}\bm{R}_{b}^{T}\bm{v} + m\bm{g},
\end{equation}

\begin{equation}
	\dot{\bm{q}} = \frac{1}{2} \bm{q} \otimes \begin{bmatrix} 0 \\ \bm{\omega}_b \end{bmatrix},
\end{equation}

\begin{equation}
	\bm{J} \dot{\bm{\omega}}_b = \bm{\tau} - \bm{\omega}_b \times \bm{J} \bm{\omega}_b - \bm{\tau}_{\text{drag}},
\end{equation}
where \( \bm{p} \in \mathbb{R}^3 \) denotes the position of the UAV's center of mass, \( \bm{v} \in \mathbb{R}^3 \) is the linear velocity, \( T \) is the total thrust, \( \bm{g} \) is gravity, and \( \bm{D} \in \mathbb{R}^{3\times3} \) is the translational drag coefficient matrix. Quaternion multiplication is denoted by \( \otimes \). The angular velocity and acceleration in the body frame are \( \bm{\omega}_b = [p, q, r]^T \) and \( \dot{\bm{\omega}}_b \), respectively. \( \bm{J} \) is the inertia matrix.  The relationship between total thrust and rotor torques is:

\begin{equation}
	\begin{bmatrix}
		T \\ \bm{\tau}
	\end{bmatrix}
	=
	\bm{M}_t \bm{u}_t,
\end{equation}
where \( \bm{u}_t = [u_0, u_1, u_2]^T \) is the individual rotor thrust vector, and each \( u_i = k_t \omega_i^2 \), where \( \omega_i \) is the angular speed of rotor \( i \), \( k_t \) is the thrust coefficient. The control allocation matrix \( \bm{M}_t \in \mathbb{R}^{4 \times 3} \) is defined as:

\begin{equation}
	\bm{M}_t = \begin{bmatrix}
		1 & 1 & 1 \\

		0 & r_{x,1} & -r_{x,2} \\
				-r_{y,0} & r_{y,1} & r_{y,2} \\
		-k_y & k_y & k_y
	\end{bmatrix},
\end{equation}
where $ r_{x,i}$ and $r_{y,i} $ are related to the horizontal distances from UAV center of gravity to the rotor.

\subsection{Control System Design}
\label{sec:nmpc_indi}

The system state is defined as \( \bm{x} = [\bm{p}^T, \bm{q}^T, \bm{v}^T, \bm{\omega}_b^T]^T \), and the control input as \( \bm{u} \in \mathbb{R}^3 \). The discrete-time model is represented as:

\begin{equation}
	\bm{x}_{k+1} = f(\bm{x}_k, \bm{u}_k).
\end{equation}
The nonlinear MPC cost function over horizon \( N \) and step size \( dt = h/N \) is formulated as:

\begin{equation}
	\begin{aligned}
		\bm{u}^*_{k:k+N-1} = \arg\min_{\bm{u}} \ & \bm{y}_{k+N}^T \bm{Q}_N \bm{y}_{k+N} + \sum_{i=k}^{k+N-1} (\bm{y}_i^T \bm{Q} \bm{y}_i + \bm{u}_i^T \bm{Q}_u \bm{u}_i) \\
		\text{s.t.} \quad & \bm{x}_{i+1} = f(\bm{x}_i, \bm{u}_i), \\
		& \underline{\bm{u}} \leq \bm{u}_i \leq \overline{\bm{u}}, 
	\end{aligned}
\end{equation}
where \( \bm{y}_i \) is the cost vector at time step \( i \), and the matrices \( \bm{Q}, \bm{Q}_N, \bm{Q}_u \) are positive-definite. $\underline{\bm{u}}$ and  $\overline{\bm{u}}$ are the lower and upper limits of $\bm{u}_i$, respectively.

A tilt-prioritized reduced-attitude control strategy is adopted to regulate the attitude control \cite{Brescianini2020Tilt}. The attitude error between the current and desired quaternions is:

\begin{equation}
	\tilde{\bm{q}} = \bm{q}_r \otimes \bm{q}^{-1} = [q_e^0, q_e^1, q_e^2, q_e^3]^T.
\end{equation}
The attitude error is further decomposed into:

\[
\tilde{\bm{q}} = \tilde{\bm{q}}_z \otimes \tilde{\bm{q}}_{xy},
\]
where \( \tilde{\bm{q}}_z \) handles yaw and \( \tilde{\bm{q}}_{xy} \) handles roll/pitch errors. The cost vector is then:

\begin{equation}
	\bm{y}_i = \begin{bmatrix}
		\bm{p} - \bm{p}_{\text{ref}} \\
		\tilde{\bm{q}}_e \\
		\bm{v} - \bm{v}_{\text{ref}} \\
		\bm{\omega}_b - \bm{\omega}_{\text{ref}} \\
		\bm{u}_t - \bm{u}_{\text{ref}}
	\end{bmatrix},
\end{equation}
where $\tilde{\bm{q}}_e$ denotes the reduced attitude error and the weight matrices are:

\begin{equation}
	\bm{Q} = \text{diag}(\bm{Q}_p, \bm{Q}_q, \bm{Q}_v, \bm{Q}_\omega, \bm{Q}_u), \quad \bm{Q}_N = \bm{Q}.
\end{equation}
The references \( \bm{x}_{\text{ref}} \) may be constant (e.g., hover) or from a desired trajectory.

Continuous high-speed rotation substantially alters UAV aerodynamics, introducing unknown and complex disturbances. To achieve precise control, we employ INDI as a low-level controller. INDI directly estimates external disturbances without requiring a full dynamic model. Following \cite{tal2020accurate}, the estimated disturbance torque is computed as:  

\begin{equation}
	\hat{\bm{\tau}}_a = -\bm{\tau}_f + \bm{J} \dot{\bm{\omega}}_f + \bm{\omega}_f \times \bm{J} \bm{\omega}_f + \bm{\tau}_{\text{drag}},
\end{equation}
where \( \bm{\tau}_f \) is the estimated torque from filtered rotor thrust, and \( \bm{\omega}_f, \dot{\bm{\omega}}_f \) are filtered angular velocity and acceleration. The desired torque is calculated as:

\begin{equation}
	\bm{\tau}_d = \bm{\tau}_f + \bm{J} \dot{\bm{\omega}}_d - \bm{J} \dot{\bm{\omega}}_f.
\end{equation}
where $\dot{\bm{\omega}}_d $ is desired angular acceleration. Finally, the INDI-based rotor thrust command is obtained as:

\begin{equation}
	\bm{u}_{\text{indi}} = \bm{M}_t^{-1}
	\begin{bmatrix}
		 T  \\
		\bm{\tau}_d
	\end{bmatrix},
\end{equation}
where $\bm{M}_t^{-1}$ denotes the Moore-Penrose inversion of $\bm{M}_t$.

\definecolor{mygray}{gray}{.9}
\begin{table}[!t]
	\renewcommand{\arraystretch}{2}
	\caption{Parameters of SPINNER Dynamics}
	\label{table_2}
	\centering
	\begin{tabular}{llll}
		\toprule
		Parameter & Value & Parameter & Value \\
		\midrule
	\rowcolor{mygray}	$m$ [$kg$] & 1.15 & 	 $\bm{M}$ [$gm^{2}$] & diag(5.59, 5.77, 6.05)\\
	 $k$ & 0.015 & $k_{t}$ [$N$] & $1.41e^{-8}$ \\
		\rowcolor{mygray}	$r_{x,1}$ [$m$] & $0.108$ &  $\bm{D}$ [$kg/s$] & diag(0.48, 0.50, 0.65)  \\
			$r_{x,2}$ [$m$] & $0.108$ &  	$r_{y,0}$ [$m$] & $0.125$  \\
			\rowcolor{mygray}	$r_{y,1}$ [$m$] & $0.063$ &  	$r_{y,2}$ [$m$] & $0.063$  \\

		\bottomrule
	\end{tabular}
\end{table}

 \begin{figure}
	\centering
	\includegraphics[width=0.950\linewidth]{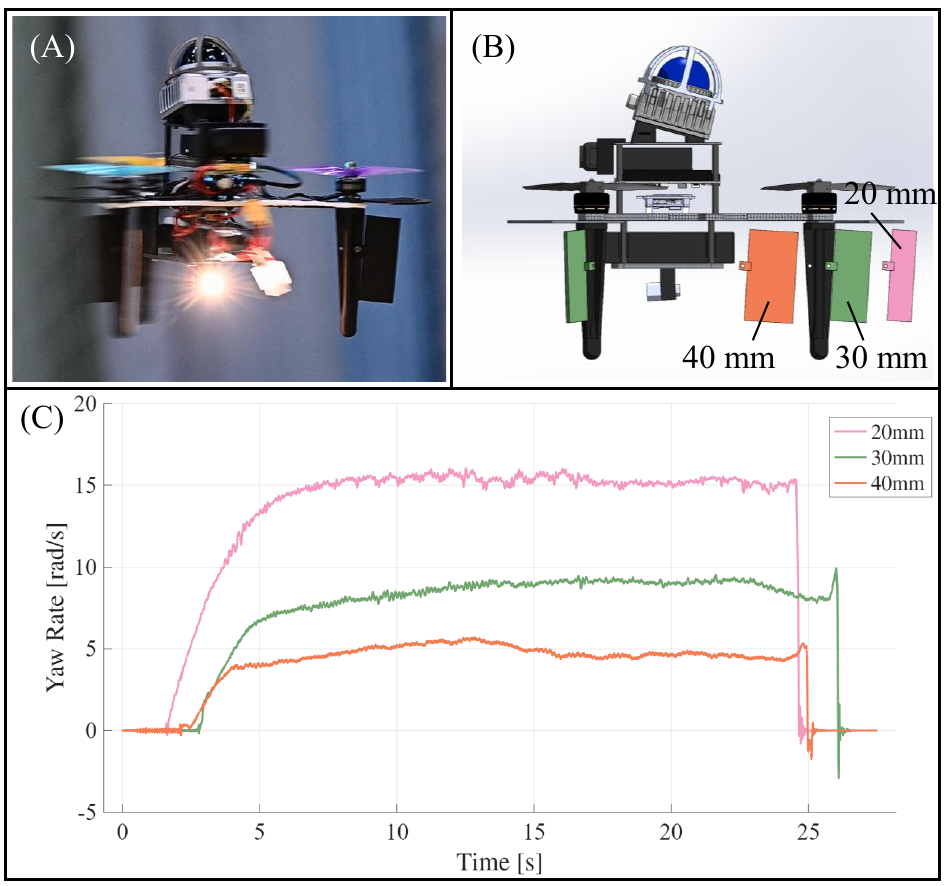}
	\caption{\textcolor{black}{    Yaw rate response of the UAV with different plate configurations. (A)  Snapshot of a real flight experiment with {SPINNER} equipped with a drag plate.
			(B) Schematic of plate configurations with a fixed length of 77\,mm and widths of 20\,mm, 30\,mm, and 40\,mm, respectively. (C) Experimental results showing the yaw rate responses for the three plate sizes.
	}}
	\label{DFB}
\end{figure}
\begin{figure*}
	\centering
	\includegraphics[width=0.850\linewidth]{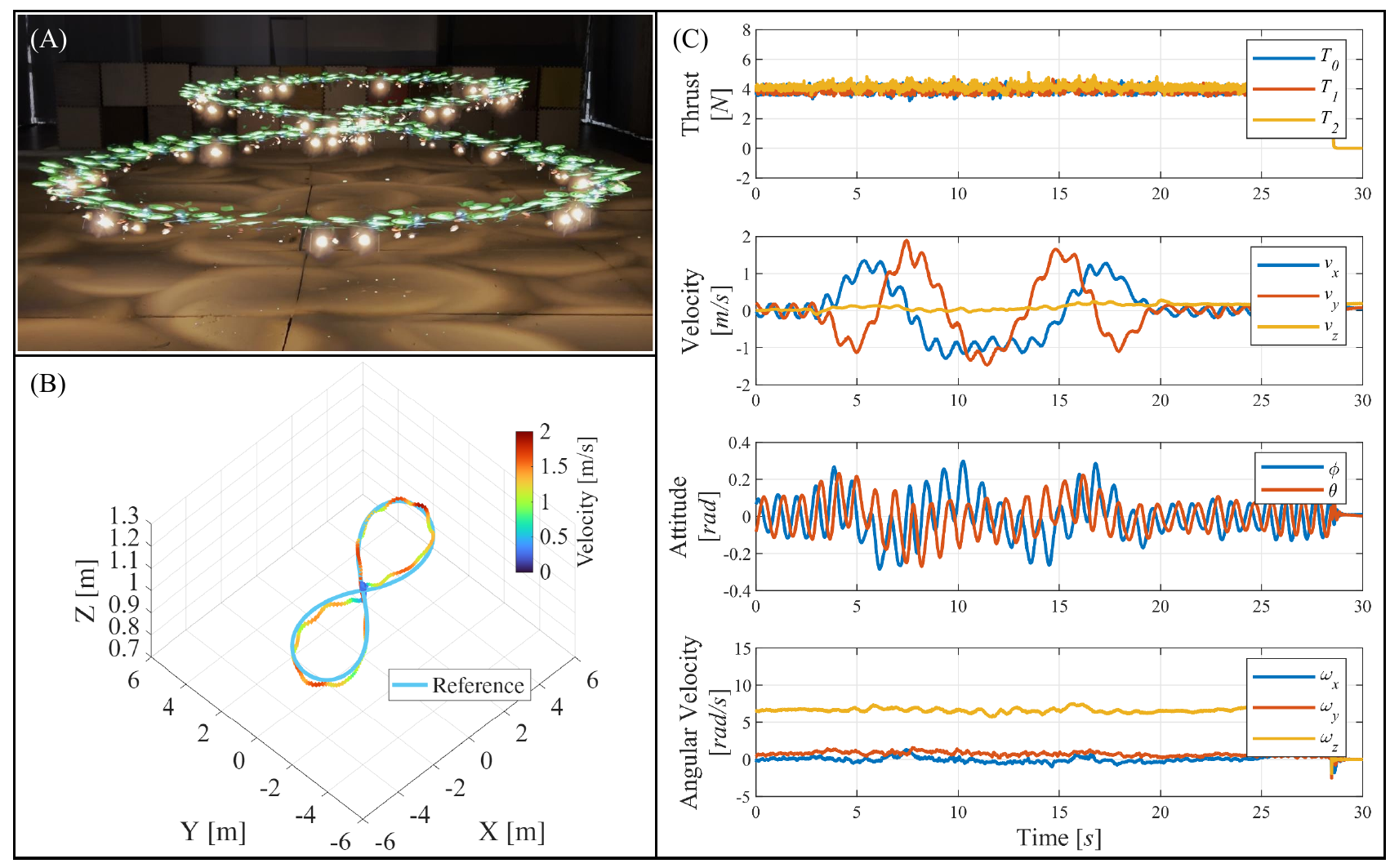}
	\caption{\textcolor{black}{ Trajectory tracking in an indoor environment. (A) Overlaid snapshots of {SPINNER} autonomously following a lemniscate trajectory. (B) Trajectory tracking performance at a maximum velocity of $2.0\,\mathrm{m/s}$. (C) Time histories of thrust and state responses under the proposed control system.
	}}
	\label{8zi}
\end{figure*}

\section{  Real-World Experiments}
\label{ Real-World Experiments}

We evaluate the effectiveness of SPINNER through extensive real-world experiments. This section addresses the following critical questions:

\begin{enumerate}
	\item How do the anti-torque plates perform in real-world flight tests? (Refer to Section~\ref{Anti-torque vanes design}.)
	\item How does the control system respond to agile trajectories and wind disturbances? (Refer to Sections~\ref{Trajectory tracking in indoor environment} and~\ref{Anti-disturbance}.)
	\item What is the FoV expansion and perception performance of \textsc{SPINNER} in real-world environments? (Refer to Sections~\ref{Indoor Navigation} and~\ref{Autonomous Navigation in Windy Wood}.)
\end{enumerate}

The physical parameters of \textsc{SPINNER} are provided in Table~\ref{table_2}. The nonlinear MPC problem is solved using the ACADO solver~\cite{houska2011acado} in conjunction with qpOASES~\cite{ferreau2014qpoases}. The prediction horizon length \(N\) and time step \(dt\) are set to 20 and 50\,ms, respectively. The parameters of nonlinear MPC are detailed in Table~\ref{table_3}.

%
 To quantitatively evaluate the tracking performance, we define two metrics: the mean tracking error  and the maximum tracking error. Let $\bm{p}_t = [x_t, y_t, z_t]^\top$ denote the UAV position at time $t$, and $\bm{p}_r = [x_r, y_r, z_r]^\top$ the corresponding reference position. The mean tracking error is given by  
\begin{equation}
	e_t =
	\sqrt{
		\frac{1}{t_2 - t_1}
		\int_{t_1}^{t_2}
		\lVert \bm{p}_t - \bm{p}_r \rVert^2 \, dt
	}
\end{equation}
 where $[t_1, t_2]$ represents the evaluation time interval. The maximum error is defined as  
  \begin{equation}
 	e_{pe} = \max \left( \lVert \bm{p}_t - \bm{p}_r \rVert \right),
 \end{equation}
  which corresponds to the maximum deviation between the UAV and the reference.

 \begin{table}[!t]
 	\renewcommand{\arraystretch}{1.5} 
 	\caption{Parameters Selection for Control System}
 	\label{table_3}
 	\centering
 	\begin{tabular}{llll}
 		\toprule
 		Parameter& Value & Parameter & Value \\
 		\midrule
 		$ \bm{Q}_{p} $ & $\mathrm{diag}(100, 100, 800)$ &	$ \bm{Q}_{v} $ & $\mathrm{diag}(1, 1, 1)$ \\
 		\rowcolor{mygray} $ \bm{Q}_{q} $ & $\mathrm{diag}(60, 60, 60, 0)$ & 	$ \bm{Q}_u $ & $\mathrm{diag}(1, 1, 1)$ \\
 		$ \bm{Q}_{\omega} $ & $\mathrm{diag}(1, 1, 0)$ &   \\
 		
 		\bottomrule
 	\end{tabular}
 \end{table}

\subsection{Anti-Torque Plate Design}
\label{Anti-torque vanes design}

To mitigate excessive spinning caused by the unbalanced yaw torque of {SPINNER}, we introduce a passive aerodynamic module by symmetrically attaching distributed anti-torque plates to the landing gear. These plates increase air resistance to counteract the net motor torque.

As shown in Fig.~\ref{DFB}(A)–(B), we test three plate configurations with a fixed length of 77\,mm and widths of 20\,mm, 30\,mm, and 40\,mm. In each trial, the UAV hovers at the starting point with three identical plates of a given width, and its angular velocity is recorded. The results in Fig.~\ref{DFB}C show average yaw rates of approximately 15.2\,rad/s, 9.3\,rad/s, and 5.4\,rad/s for the 20\,mm, 30\,mm, and 40\,mm plates, respectively. High spin rates accelerate the elimination of FoV blind zones, but overly fast updates can interrupt the convergence of LiDAR odometry, potentially leading to divergence. Meanwhile, larger plates increase aerodynamic drag, which degrades trajectory tracking performance. Considering this trade-off, we adopt the 30\,mm plate configuration as the standard setup for subsequent experiments.

\begin{figure}
	\centering
	\includegraphics[width=0.950\linewidth]{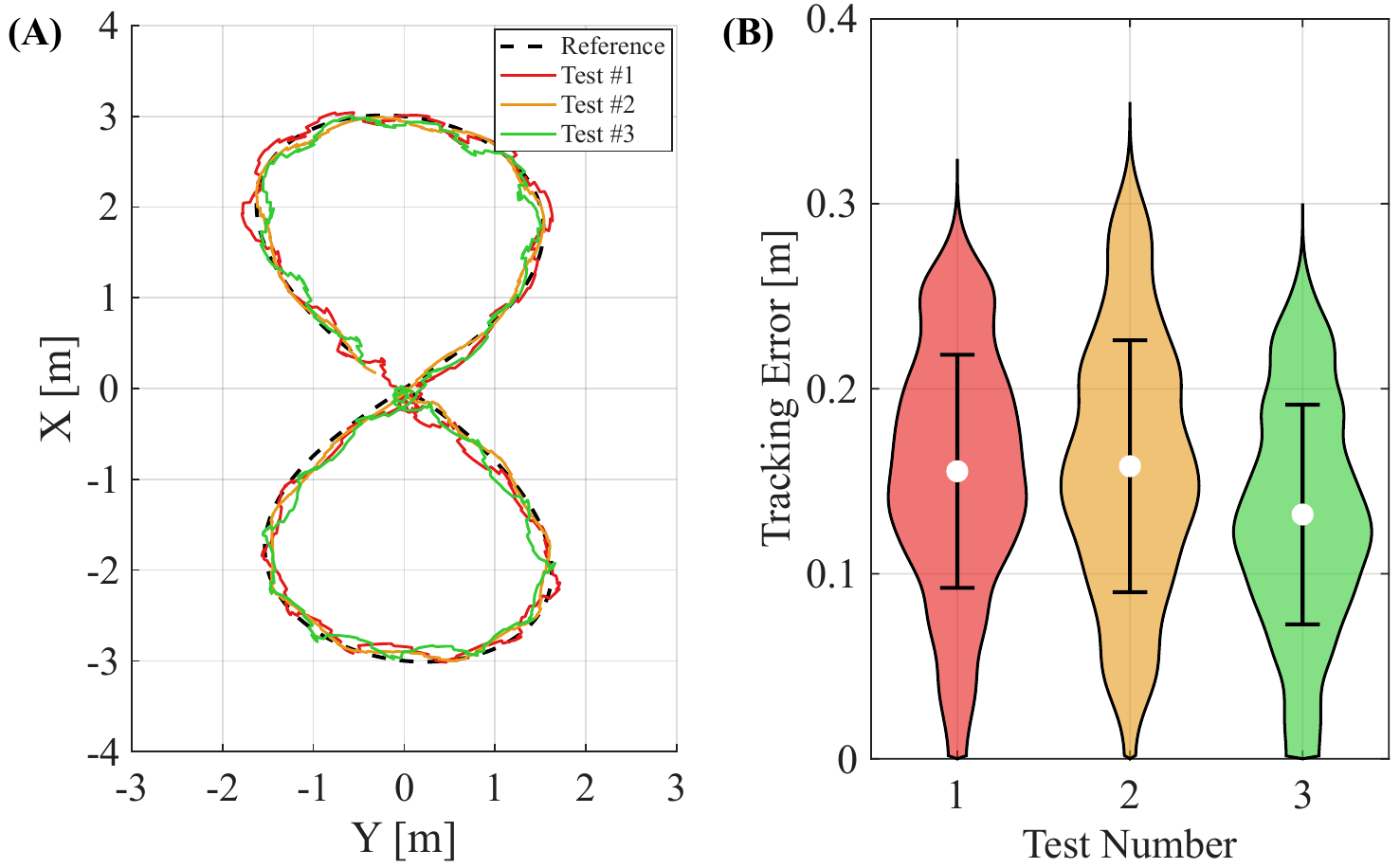}
	\caption{\textcolor{black}{  Experimental results from three trajectory tracking tests. (A) Comparison between the reference trajectory and real-world flight trajectories. (B) Violin plots showing the distribution of position-tracking errors across the three tests. The shape of each violin indicates the data density, the white point marks the mean, and the black bar represents the standard deviation.  }}
	\label{traj_violin}
\end{figure}

\begin{figure}
	\centering
	\includegraphics[width=0.99\linewidth]{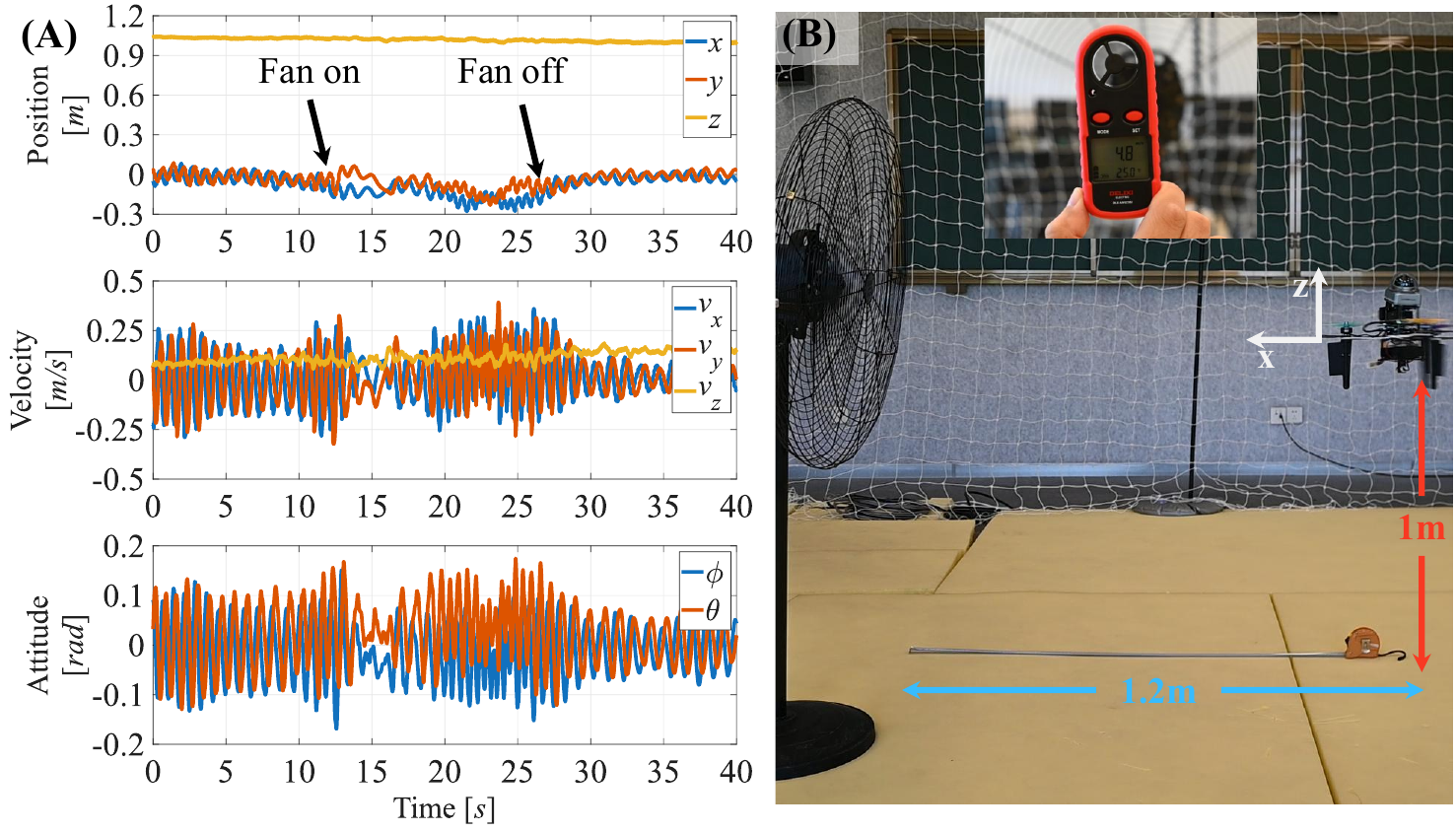}
	\caption{\textcolor{black}{ External disturbance rejection test during UAV hovering. (A) Responses of position, attitude angles, and yaw rate with and without wind disturbance. (B) Experimental setup: the UAV hovers 1.2\,m from the fan at a height of 1\,m, where the fan generates a wind gust of 4.8\,m/s at the hovering position of {SPINNER}.  }}
	\label{antidisturbance}
\end{figure}

\subsection{Trajectory Tracking Performance }
\label{Trajectory tracking in indoor environment}

To evaluate the tracking performance of {SPINNER}, we conduct high-speed trajectory-tracking experiments in an indoor environment (see Fig.~\ref{8zi}(A)). The reference path is a lemniscate trajectory with a maximum target velocity of $2.0\,\mathrm{m/s}$. This test assesses the UAV's dynamic response under large maneuvering demands. As shown in Fig.~\ref{8zi}(B), the lemniscate trajectory extends along all three spatial directions to fully excite the system dynamics: $6.0\,\mathrm{m}$ along the $x$-axis, $3.0\,\mathrm{m}$ along the $y$-axis, and $0.6\,\mathrm{m}$ along the $z$-axis. Figure~\ref{8zi}(C) illustrates the trajectory-tracking and thrust-response results. Three tracking trials are presented in Fig.~\ref{traj_violin}, with mean tracking errors of $0.15\,\mathrm{m}$, $0.16\,\mathrm{m}$, and $0.13\,\mathrm{m}$, respectively. Overall, {SPINNER} closely follows the desired path with minimal error, demonstrating strong tracking capability.

\subsection{Anti-Disturbance Performance }
\label{Anti-disturbance}

UAVs are often affected by external disturbances such as wind gusts. To evaluate the robustness of the proposed tri-rotor UAV under such conditions, we conduct a disturbance rejection experiment using an industrial fan to generate gusts and measure the UAV’s displacement from its initial hover position. As shown in Fig.~\ref{antidisturbance}, the UAV hovers at a height of $1.0\,\mathrm{m}$ and $1.2\,\mathrm{m}$ in front of the fan, which produces gusts with wind speeds up to $4.8\,\mathrm{m/s}$. Once exposed to the wind, the UAV is displaced from its original position with a maximum error of approximately $0.31\,\mathrm{m}$ and a mean error of $0.12\,\mathrm{m}$. The onboard controller detects the deviation and compensates for the disturbance, gradually returning the UAV to its original position within $12\,\mathrm{s}$. During this recovery process, coupling effects inherent to spinning flight cause minor deviations along the $y$- and $z$-axes, although the primary disturbance acts along the $x$-axis. In addition, significant oscillations occur in the pitch and roll angles, while the yaw rate exhibits larger variations to modulate motor thrust distribution. This experiment demonstrates that the proposed UAV maintains stable hovering and attitude control under external disturbances, confirming its suitability for real-world operational scenarios.

 \begin{figure}
	\centering
	\includegraphics[width=0.850\linewidth]{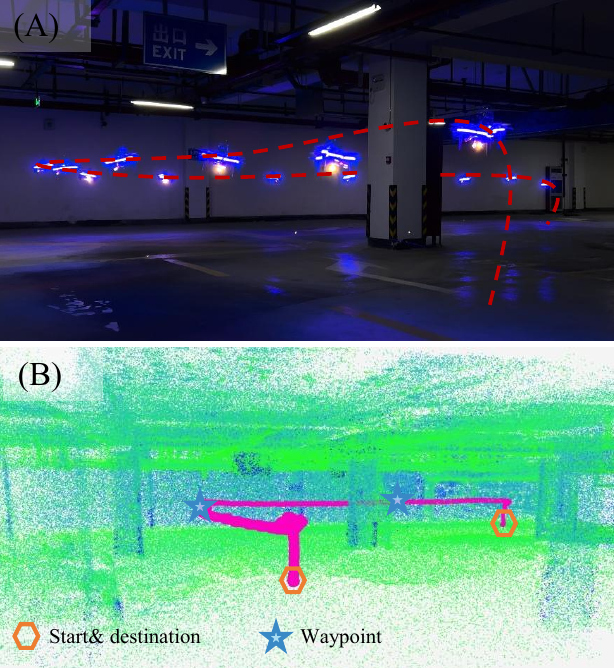}
	\caption{\textcolor{black}{  (A)–(B) Real flight process and the corresponding 3D point cloud map in the parking lot, with the flight trajectory shown as a red path.
	  }}
	\label{cheku}
\end{figure}

\subsection{Indoor Navigation}
\label{Indoor Navigation}

To evaluate the fully autonomous navigation capability of the proposed UAV in unknown environments, we conduct a waypoint-based navigation experiment in an underground parking lot of over $1000\,\mathrm{m}^2$ (see Fig.~\ref{cheku}A–B). During the experiment, the UAV operates without external positioning systems or prior environmental maps. A set of sparse waypoints is provided, and the onboard trajectory planner computes smooth paths between them in real time. As illustrated in Fig.~\ref{cheku}A, the planner continuously generates smooth reference trajectories (red curve) connecting the UAV’s current position to the next waypoint. These trajectories are accurately tracked by the onboard controller, as evidenced by the actual flight path shown in red in Fig.~\ref{cheku}B. The UAV completes the $63\,\mathrm{m}$ path in  $125\,\mathrm{s}$. Throughout the process, the vehicle operates fully autonomously without manual intervention or remote control. Meanwhile, a dense 3D point cloud map of the environment is built onboard. Owing to the FoV expansion provided by the spinning motion, the resulting point cloud is evenly distributed across the horizontal plane rather than constrained within a narrow sensor cone, demonstrating improved environmental coverage and exploration efficiency.

\subsection{Autonomous Navigation in Windy Forest}
\label{Autonomous Navigation in Windy Wood}

We conduct field experiments in a forested outdoor environment, as shown in Fig.~\ref{forest}. Multiple successful flights demonstrate the UAV’s strong navigation capability in unknown, GNSS-denied conditions. These results confirm that the proposed platform performs agile maneuvers while simultaneously perceiving the surrounding environment. With body rotation-induced FoV expansion, the UAV achieves perception coverage well beyond the native vertical FoV of its onboard sensors in all directions, significantly improving environmental awareness and adaptability in complex natural scenarios.

\section{Conclusion }
\label{Conclusion}

This article presented {SPINNER}, a self-rotating tri-rotor UAV designed for FoV expansion and autonomous flight. The unique merits of the proposed UAV include the following:  
1) The self-rotating design expands the effective FoV of onboard sensors without extra hardware, while passive anti-torque plates counter motor torque to maintain a controllable rotation speed; 
2) A nonlinear control framework that combines nonlinear MPC and INDI effectively compensates for the strong nonlinearities and couplings induced by spinning flight.  
Comprehensive experiments demonstrate that {SPINNER} maintains stable flight under wind disturbances up to $4.8\,\mathrm{m/s}$, achieves accurate trajectory tracking at speeds up to $2.0\,\mathrm{m/s}$, and performs robust autonomous navigation in GNSS-denied environments such as parking garages and forested areas. These results confirm the effectiveness of the proposed UAV in enhancing perception efficiency and adaptability in complex real-world scenarios.

\ifCLASSOPTIONcaptionsoff
\newpage
\fi

\bibliographystyle{IEEEtran}
\bibliography{IEEEabrv,mylib}

\begin{thebibliography}{10}
\providecommand{\url}[1]{#1}
\csname url@samestyle\endcsname
\providecommand{\newblock}{\relax}
\providecommand{\bibinfo}[2]{#2}
\providecommand{\BIBentrySTDinterwordspacing}{\spaceskip=0pt\relax}
\providecommand{\BIBentryALTinterwordstretchfactor}{4}
\providecommand{\BIBentryALTinterwordspacing}{\spaceskip=\fontdimen2\font plus
\BIBentryALTinterwordstretchfactor\fontdimen3\font minus \fontdimen4\font\relax}
\providecommand{\BIBforeignlanguage}[2]{{%
\expandafter\ifx\csname l@#1\endcsname\relax
\typeout{** WARNING: IEEEtran.bst: No hyphenation pattern has been}%
\typeout{** loaded for the language `#1'. Using the pattern for}%
\typeout{** the default language instead.}%
\else
\language=\csname l@#1\endcsname
\fi
#2}}
\providecommand{\BIBdecl}{\relax}
\BIBdecl

\bibitem{mellinger2011minimum}
D.~Mellinger and V.~Kumar, ``Minimum snap trajectory generation and control for quadrotors,'' in \emph{2011 IEEE International Conference on Robotics and Automation (ICRA)}, Shanghai, China, 2011, pp. 2520--2525.

\bibitem{Philipp2022Agilicious}
P.~Foehn, E.~Kaufmann, A.~Romero, R.~Penicka, S.~Sun, L.~Bauersfeld, T.~Laengle, G.~Cioffi, Y.~Song, A.~Loquercio, and D.~Scaramuzza, ``Agilicious: Open-source and open-hardware agile quadrotor for vision-based flight,'' \emph{Science Robotics}, vol.~7, no.~67, p. eabl6259, 2022.

\bibitem{Wang2024Multi}
D.~Wang, J.~Wang, S.~He, J.~Huang, B.~Zhang, Y.~Mao, G.~Huang, C.~Xu, and F.~Gao, ``Multi-fov-constrained trajectory planning for multirotor safe landing,'' in \emph{2024 IEEE/RSJ International Conference on Intelligent Robots and Systems (IROS)}, 2024, pp. 5356--5363.

\bibitem{chen2023self}
N.~Chen, F.~Z. Kong, W.~Xu, Y.~X. Cai, H.~T. Li, D.~J. He, Y.~M. Qin, and F.~Zhang, ``A self-rotating, single-actuated {UAV} with extended sensor field of view for autonomous navigation,'' \emph{Science Robotics}, vol.~8, no.~76, p. eade4538, 2023.

\bibitem{Gurtner2009Investigation}
A.~Gurtner, D.~G. Greer, R.~Glassock, L.~Mejias, R.~A. Walker, and W.~W. Boles, ``Investigation of fish-eye lenses for small-uav aerial photography,'' \emph{IEEE Transactions on Geoscience and Remote Sensing}, vol.~47, no.~3, pp. 709--721, 2009.

\bibitem{Gao2020Autonomous}
W.~Gao, K.~Wang, W.~Ding, F.~Gao, T.~Qin, and S.~Shen, ``Autonomous aerial robot using dual-fisheye cameras,'' \emph{Journal of Field Robotics}, vol.~37, no.~4, pp. 497--514, 2020.

\bibitem{Karimi2021LoLa}
M.~Karimi, M.~Oelsch, O.~Stengel, E.~Babaians, and E.~Steinbach, ``Lola-slam: Low-latency lidar slam using continuous scan slicing,'' \emph{IEEE Robotics and Automation Letters}, vol.~6, no.~2, pp. 2248--2255, 2021.

\bibitem{Liu2024OmniNxt}
P.~Liu, C.~Feng, Y.~Xu, Y.~Ning, H.~Xu, and S.~Shen, ``Omninxt: A fully open-source and compact aerial robot with omnidirectional visual perception,'' in \emph{2024 IEEE/RSJ International Conference on Intelligent Robots and Systems (IROS)}, 2024, pp. 10\,605--10\,612.

\bibitem{Nan2022Nonlinear}
F.~Nan, S.~H. Sun, P.~Foehn, and D.~Scaramuzza, ``Nonlinear {{MPC}} for quadrotor fault-tolerant control,'' \emph{IEEE Robotics and Automation Letters}, vol.~7, no.~2, pp. 5047--5054, 2022.

\bibitem{zhou2024internal}
X.~Zhou, M.~Wang, C.~Cui, Y.~Wang, C.~Xu, and F.~Gao, ``Internal and external disturbances aware motion planning and control for quadrotors,'' \emph{IET Cyber-Systems and Robotics}, vol.~6, no.~3, p. e12122, 2024.

\bibitem{zhou2025rotor}
X.~Zhou, M.~Wang, C.~Li, C.~Cui, R.~Zhang, Y.~Wang, C.~Xu, and F.~Gao, ``Rotor-failure-aware quadrotors flight in unknown environments,'' \emph{arXiv preprint arXiv:2510.11306}, 2025.

\bibitem{Sun2021Autonomous}
S.~H. Sun, G.~Cioffi, C.~de~Visser, and D.~Scaramuzza, ``Autonomous quadrotor flight despite rotor failure with onboard vision sensors: Frames vs. events,'' \emph{IEEE Robotics and Automation Letters}, vol.~6, no.~2, pp. 580--587, 2021.

\bibitem{he2023point}
D.~He, W.~Xu, N.~Chen, F.~Kong, C.~Yuan, and F.~Zhang, ``Point-lio: robust high-bandwidth light detection and ranging inertial odometry,'' \emph{Advanced Intelligent Systems}, vol.~5, no.~7, p. 2200459, 2023.

\bibitem{bhardwaj2022design}
H.~Bhardwaj, X.~Cai, S.~K.~H. Win, and S.~Foong, ``Design, modeling and control of a two flight mode capable single wing rotorcraft with mid-air transition ability,'' \emph{IEEE Robotics and Automation Letters}, vol.~7, no.~4, pp. 11\,720--11\,727, 2022.

\bibitem{cai2023self}
X.~Cai, L.~S.~T. Win, B.~L. Suhadi, H.~Bhardwaj, and S.~Foong, ``A self-rotary aerial robot with passive compliant variable-pitch wings,'' \emph{IEEE Robotics and Automation Letters}, vol.~8, no.~7, pp. 4195--4202, 2023.

\bibitem{cai2023modeling}
X.~Cai, S.~K.~H. Win, H.~Bhardwaj, and S.~Foong, ``Modeling, control and implementation of adaptive reconfigurable rotary wings (arrows),'' \emph{IEEE/ASME Transactions on Mechatronics}, vol.~28, no.~4, pp. 2282--2292, 2023.

\bibitem{zhang2016controllable}
W.~Zhang, M.~W. Mueller, and R.~D'Andrea, ``A controllable flying vehicle with a single moving part,'' in \emph{2016 IEEE International Conference on Robotics and Automation (ICRA)}.\hskip 1em plus 0.5em minus 0.4em\relax IEEE, 2016, pp. 3275--3281.

\bibitem{xu2025aerial}
J.~Xu and P.~Chirarattananon, ``An aerial robot achieving bimodal flight and independent attitude control through passive morphing,'' \emph{Advanced Intelligent Systems}, vol.~7, no.~7, p. 2400464, 2025.

\bibitem{win2021agile}
S.~K.~H. Win, L.~S.~T. Win, D.~Sufiyan, G.~S. Soh, and S.~Foong, ``An agile samara-inspired single-actuator aerial robot capable of autorotation and diving,'' \emph{IEEE Transactions on Robotics}, vol.~38, no.~2, pp. 1033--1046, 2021.

\bibitem{sharp2016micro}
D.~Sharp, C.~Stoneking, and K.~Fregene, ``Micro air vehicle based navigation aiding in degraded environments,'' in \emph{2016 IEEE/ION Position, Location and Navigation Symposium (PLANS)}.\hskip 1em plus 0.5em minus 0.4em\relax IEEE, 2016, pp. 305--312.

\bibitem{isaacs2014gps}
J.~T. Isaacs, C.~Magee, A.~Subbaraman, F.~Quitin, K.~Fregene, A.~R. Teel, U.~Madhow, and J.~P. Hespanha, ``Gps-optimal micro air vehicle navigation in degraded environments,'' in \emph{2014 American Control Conference}.\hskip 1em plus 0.5em minus 0.4em\relax IEEE, 2014, pp. 1864--1871.

\bibitem{fregene2010development}
K.~Fregene, S.~Jameson, D.~Sharp, H.~Youngren, and D.~Stuart, ``Development and flight validation of an autonomous mono-wing uas,'' in \emph{American Helicopter Society Forum, Phoenix, AZ}, 2010.

\bibitem{Mavros2024git}
Mavros, https://github.com/mavlink/mavros.

\bibitem{Zheng2025FAST}
C.~Zheng, W.~Xu, Z.~Zou, T.~Hua, C.~Yuan, D.~He, B.~Zhou, Z.~Liu, J.~Lin, F.~Zhu, Y.~Ren, R.~Wang, F.~Meng, and F.~Zhang, ``Fast-livo2: Fast, direct lidar–inertial–visual odometry,'' \emph{IEEE Transactions on Robotics}, vol.~41, pp. 326--346, 2025.

\bibitem{Wang2022Zhe}
Z.~P. Wang, X.~Zhou, C.~Xu, and F.~Gao, ``Geometrically constrained trajectory optimization for multicopters,'' \emph{IEEE Transactions on Robotics}, vol.~38, no.~5, pp. 3259--3278, 2022.

\bibitem{Brescianini2020Tilt}
D.~Brescianini and R.~D’Andrea, ``Tilt-prioritized quadrocopter attitude control,'' \emph{IEEE Transactions on Control Systems Technology}, vol.~28, no.~2, pp. 376--387, 2020.

\bibitem{tal2020accurate}
E.~Tal and S.~Karaman, ``Accurate tracking of aggressive quadrotor trajectories using incremental nonlinear dynamic inversion and differential flatness,'' \emph{IEEE Transactions on Control Systems Technology}, vol.~29, no.~3, pp. 1203--1218, 2020.

\bibitem{houska2011acado}
B.~Houska, H.~J. Ferreau, and M.~Diehl, ``Acado toolkit—{An} open-source framework for automatic control and dynamic optimization,'' \emph{Optimal control applications and methods}, vol.~32, no.~3, pp. 298--312, 2011.

\bibitem{ferreau2014qpoases}
H.~J. Ferreau, C.~Kirches, A.~Potschka, H.~G. Bock, and M.~Diehl, ``{QPOASES}: A parametric active-set algorithm for quadratic programming,'' \emph{Mathematical Programming Computation}, vol.~6, pp. 327--363, 2014.

\end{thebibliography}
\end{document}